\documentclass{article}

\usepackage{microtype}
\usepackage{graphicx}
\usepackage{subfigure}
\usepackage{booktabs} 

\usepackage{hyperref}

\usepackage[accepted]{icml2025}

\usepackage{amsmath}
\usepackage{amssymb}
\usepackage{mathtools}
\usepackage{amsthm}

\usepackage{booktabs} 
\usepackage{multirow} 
\usepackage{array}    

\usepackage{amsfonts}

\def\b{\ensuremath\boldsymbol}

\usepackage[capitalize,noabbrev]{cleveref}

\theoremstyle{plain}

\theoremstyle{definition}

\theoremstyle{remark}

\usepackage[textsize=tiny]{todonotes}

\icmltitlerunning{Self-Supervised Learning by Curvature Alignment}

\begin{document}

\AddToShipoutPictureBG*{%
  \AtPageUpperLeft{%
    \setlength\unitlength{1in}%
    \hspace*{\dimexpr0.5\paperwidth\relax}
    \makebox(0,-0.75)[c]{\normalsize {\color{black}  A shorter version of this paper has been published in:} }
    }}

\AddToShipoutPictureBG*{%
  \AtPageUpperLeft{%
    \setlength\unitlength{1in}%
    \hspace*{\dimexpr0.5\paperwidth\relax}
    \makebox(0,-1.2)[c]{\normalsize {\color{black} Journal of Computational Vision and Imaging Systems, Vol. 11, No. 1, Special Issue: Proceedings of CVIS 2025. } }
    }}


\twocolumn[
\icmltitle{Self-Supervised Learning by Curvature Alignment}



\icmlsetsymbol{equal}{*}

\begin{icmlauthorlist}
\icmlauthor{Benyamin Ghojogh}{aaa}
\icmlauthor{M.Hadi Sepanj}{yyy}
\icmlauthor{Paul Fieguth}{yyy}
\end{icmlauthorlist}

\icmlaffiliation{yyy}{Vision and Image Processing Group, Systems Design Engineering, University of Waterloo, Ontario, Canada}
\icmlaffiliation{aaa}{Artificial Intelligence Scientist, Waterloo, Ontario, Canada}

\icmlcorrespondingauthor{Benyamin Ghojogh}{bghojogh@uwaterloo.ca}
\icmlcorrespondingauthor{M.Hadi Sepanj}{mhsepanj@uwaterloo.ca}
\icmlcorrespondingauthor{Paul Fieguth}{paul.fieguth@uwaterloo.ca}

\icmlkeywords{Machine Learning}

\vskip 0.3in
]



\printAffiliationsAndNotice{Benyamin Ghojogh and M.Hadi Sepanj contributed equally to this work.}  

\begin{abstract}
Self-supervised learning (SSL) has recently advanced through non-contrastive methods that couple an invariance term with variance, covariance, or redundancy-reduction penalties. While such objectives shape first- and second-order statistics of the representation, they largely ignore the local geometry of the underlying data manifold. In this paper, we introduce \emph{CurvSSL}, a curvature-regularized self-supervised learning framework, and its RKHS extension, \emph{kernel CurvSSL}. Our approach retains a standard two-view encoder--projector architecture with a Barlow Twins-style redundancy-reduction loss on projected features, but augments it with a curvature-based regularizer. Each embedding is treated as a vertex whose $k$ nearest neighbors define a discrete curvature score via cosine interactions on the unit hypersphere; in the kernel variant, curvature is computed from a normalized local Gram matrix in an RKHS. These scores are aligned and decorrelated across augmentations by a Barlow-style loss on a curvature-derived matrix, encouraging both view invariance and consistency of local manifold bending. Experiments on MNIST and CIFAR-10 datasets with a ResNet-18 backbone show that curvature-regularized SSL yields competitive or improved linear evaluation performance compared to Barlow Twins and VICReg. Our results indicate that explicitly shaping local geometry is a simple and effective complement to purely statistical SSL regularizers.
\end{abstract}

\section{Introduction}

Self-supervised learning (SSL) has become a central paradigm for visual representation learning, replacing explicit labels with surrogate objectives defined over augmented views of the same image~\cite{sepanj2025ieeeaccess,rani2023self}. Contrastive methods such as InfoNCE-based approaches~\cite{sepanj2025sinsim,chen2020simple} maximize agreement between positive pairs while repelling negatives, whereas recent non-contrastive methods~\cite{grill2020bootstrap,bardes2022vicreg,sepanj2024aligning,sepanj2025kernel} avoid explicit negatives by combining an invariance term with variance, covariance, or redundancy-reduction penalties. Architectures such as Barlow Twins \cite{zbontar2021barlow} and VICReg \cite{bardes2022vicreg} enforce that two augmentations of the same sample produce highly correlated embeddings along the diagonal of a cross-correlation matrix, while off-diagonal terms and per-dimension variances are controlled to prevent dimensional collapse.

These approaches, however, still treat the representation space primarily as a flat Euclidean vector space and regularize it using first- and second-order statistics (means, variances, cross-correlations). From a geometric point of view, high-dimensional data are often assumed to concentrate around a lower-dimensional manifold embedded in feature space. Standard SSL objectives encourage different augmentations of the same input to map to nearby points on this manifold and to occupy decorrelated feature dimensions globally, but they do not explicitly control the \emph{local} geometry of the learned manifold. In particular, two augmentations may be close in Euclidean distance yet induce different local neighborhoods, tangent directions, or higher-order bending of the manifold. As a consequence, embeddings can satisfy invariance and redundancy-reduction constraints while still distorting the local structure that underlies nearest-neighbor retrieval, clustering, or semi-supervised learning.

In differential and discrete geometry, curvature quantifies how a surface or manifold bends in a neighborhood of a point. For polyhedra, classical constructions based on angular defect measure how much the sum of face angles around a vertex deviates from $2\pi$~\cite{descartes1890progymnasmata,markvorsen1996curvature,coxeter1973regular,richeson2019euler,hilton1982descartes}. The sharper the corner, the larger the defect. A closely related discrete viewpoint for data is to imagine each point as a vertex of a hypothetical polyhedron whose faces are spanned by its $k$ nearest neighbors \cite{ghojogh2020anomaly}. By translating neighbor differences to the origin, normalizing them onto a unit hypersphere, and aggregating cosine similarities between neighbor directions, one obtains a scalar \emph{curvature score} that reflects how sharply the local neighborhood bends around that point. This construction can be further generalized to reproducing kernel Hilbert spaces (RKHS)~\cite{gretton2013introduction,sepanj2025kernel} by expressing inner products and norms through a kernel function, and normalizing the corresponding local Gram matrix.

Motivated by this geometric perspective, we propose
\emph{curvature-regularized self-supervised learning (CurvSSL)} along with its kernel version \textit{kernel CurvSSL}.
This method is a simple non-contrastive SSL objective that integrates curvature into the learning signal.
We retain a standard two-view encoder--projector architecture and a redundancy-reduction term in the spirit of Barlow Twins \cite{zbontar2021barlow}, which encourages diagonal cross-correlation between the projected embeddings of two augmentations while driving off-diagonal correlations toward zero. On top of this, we treat each projected embedding as a vertex in representation space, compute a discrete curvature score from its $k$ nearest neighbors via cosine interactions on the unit hypersphere, and use these scores to define an additional, geometry-aware regularizer. At the batch level, we align curvature across augmentations of the same samples and decorrelate curvature patterns across different samples using a Barlow Twins-style loss on a curvature-derived matrix. In this way, the objective does not only enforce invariance and low redundancy in the coordinates of the embeddings, but also promotes consistency and diversity in the local bending of the learned manifold.

Overall, CurvSSL can be viewed as a curvature-regularized variant of non-contrastive SSL: the backbone loss still enforces view-invariance and redundancy reduction, yet the representation is further constrained to preserve local manifold geometry as captured by discrete curvature in the embedding space (and, in an extension, kernelized curvature in an RKHS). We show experimentally that this simple curvature-aware modification of a ResNet-based SSL pipeline yields competitive representations, indicating that explicitly shaping local geometry is a promising complement to purely statistical regularizers.

\section{Background on Polyhedron Curvature and Angular Defect}

\begin{figure}[!t]
\centering
\includegraphics[width=3.2in]{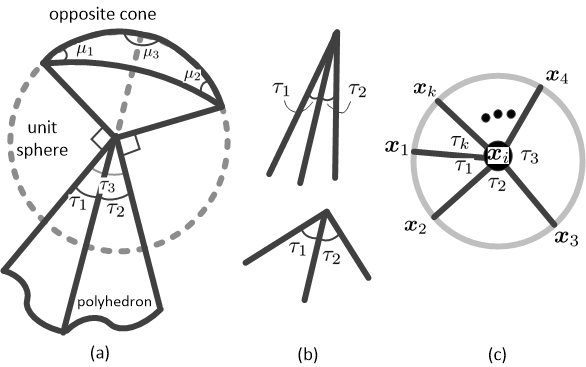}
\caption{(a) Polyhedron vertex, unit sphere, and the opposite cone, (b) large and small curvature, (c) a point and its neighbors normalized on a unit hyper-sphere around it.}
\label{figure_polyhedron}
\end{figure}

A \textit{polytope} is a geometrical object in $\mathbb{R}^d$ whose faces are planar. The special cases of polytope in $\mathbb{R}^2$ and $\mathbb{R}^3$ are called \textit{polygon} and \textit{polyhedron}, respectively. 
Some examples for polyhedron are cube, tetrahedron, octahedron, icosahedron, and dodecahedron with four, eight, and twenty triangular faces, and twelve flat faces, respectively \cite{coxeter1973regular}.
Consider a polygon where $\tau_j$ and $\mu_j$ are the interior and exterior angles at the $j$-th vertex; we have $\tau_j + \mu_j = \pi$.
A similar analysis holds in $\mathbb{R}^3$ for Fig. \ref{figure_polyhedron}-a. In this figure, a vertex of a polyhedron and its opposite cone are shown where the opposite cone is defined to have perpendicular faces to the faces of the polyhedron at the vertex. The intersection of a unit sphere centered at the vertex and the opposite cone is shown in the figure.
This intersection is a geodesic on the unit sphere.
According to Thomas Harriot's theorem proposed in 1603 \cite{markvorsen1996curvature}, if this geodesic on the unit sphere is a triangle, its area is $\mu_1 + \mu_2 + \mu_3 - \pi = 2\pi - (\tau_1 + \tau_2 + \tau_3)$. 
The generalization of this theorem from a geodesic triangular polygon ($3$-gon) to an $k$-gon is \cite{markvorsen1996curvature}:
\begin{align}
\mu_1 + \dots + \mu_k - k \pi + 2\pi = 2\pi - \sum_{a=1}^k \tau_a,
\end{align}
where the polyhedron has $k$ faces meeting at the vertex. 

Ren{\'e} Descartes's \textit{angular defect} at a vertex $\b{x}$ of a polyhedron is \cite{descartes1890progymnasmata}: 
\begin{align}
\mathcal{D}(\b{x}) := 2\pi - \sum_{a=1}^k \tau_a.
\end{align}
The total defect of a polyhedron is defined as the summation of the defects over the vertices.
It can be shown that the total defect of a polyhedron with $v$ vertices, $e$ edges, and $f$ faces is: 
\begin{align}
\mathcal{D} := \sum_{i=1}^v \mathcal{D}(\b{x}_i) = 2\pi (v - e + f).
\end{align}
The term $v - e + f$ is Euler-Poincar{\'e} characteristic of the polyhedron \cite{richeson2019euler,hilton1982descartes}; therefore, the total defect of a polyhedron is equal to its Euler-Poincar{\'e} characteristic. 
According to Fig. \ref{figure_polyhedron}-b, the smaller $\tau$ angles result in sharper corner of the polyhedron. Therefore, we can consider the angular defect as the \textit{curvature} of the vertex. 

\section{Curvature Calculation for Data Points}

\subsection{Curvature Calculation in the Input Space}

The main idea of the \textit{curvature calculation of data points} is as follows \cite{ghojogh2020anomaly}. 
Every data point is considered to be the vertex of a hypothetical polyhedron (see Fig. \ref{figure_polyhedron}-a). 
For every point, we find its $k$-Nearest Neighbors ($k$-NN).
The $k$ neighbors of the point (vertex) form the $k$ faces of a polyhedron meeting at that vertex. 
Then, the more curvature that point (vertex) has, the more anomalous it is because it is far away (different) from its neighbors. 
Therefore, we define a \textit{curvature score}, denoted by $c$, which is proportional to the curvature or angular effect. 

Since, according to the equation of angular effect, the curvature is proportional to negative summation of angles, we can consider the curvature score to be inversely proportional to the summation of angles. Without loss of generality, we assume the angles are in the range $[0, \pi]$ (otherwise, we take the smaller angle). The less the angles between two edges of the polyhedron, the more their cosine. 
As the curvature score is inversely proportional to the angles, we can use cosine for the curvature score: 
\begin{align}
c(\b{x}_i) \propto \frac{1}{\tau_a} \propto \cos(\tau_a).
\end{align}
We define the curvature score to be the summation of cosine of the angles of the polyhedron faces meeting at that point:
\begin{align}
c(\b{x}_i) := \sum_{a=1}^{k} \cos(\tau_a) = \sum_{a=1}^{k} \frac{\breve{\b{x}}_a^\top \breve{\b{x}}_{a+1}}{||\breve{\b{x}}_a||_2 ||\breve{\b{x}}_{a+1}||_2},
\end{align}
where $\breve{\b{x}}_a := \b{x}_a - \b{x}_i$ is the $a$-th edge of the polyhedron passing through the vertex $\b{x}_i$, $\b{x}_a$ is the $a$-th neighbor of $\b{x}_i$, and $\breve{\b{x}}_{a+1}$ denotes the next edge sharing the same polyhedron face with $\breve{\b{x}}_a$ where $\breve{\b{x}}_{k+1}=\breve{\b{x}}_1$. 

Note that finding the pairs of edges which belong to the same face is difficult and time-consuming so we relax this calculation to the summation of the cosine of angles between all pairs of edges meeting at the vertex $\b{x}_i$:
\begin{align}\label{equation_curvature_score}
c(\b{x}_i) := \sum_{a=1}^{k-1} \sum_{b=a+1}^k \frac{\breve{\b{x}}_a^\top \breve{\b{x}}_b}{||\breve{\b{x}}_a||_2 ||\breve{\b{x}}_b||_2},
\end{align}
where $\breve{\b{x}}_a := \b{x}_a - \b{x}_i$, $\breve{\b{x}}_b := \b{x}_b - \b{x}_i$, and $\b{x}_a$ and $\b{x}_b$ denote the $a$-th and $b$-th neighbors of $\b{x}_i$. 
In Eq. (\ref{equation_curvature_score}), we have omitted the redundant angles because of symmetry of inner product.
Note that the Eq. (\ref{equation_curvature_score}) implies that we normalize the $k$ neighbors of $\b{x}_i$ to fall on the unit hyper-sphere centered at $\b{x}_i$ and then compute their cosine similarities (see Fig. \ref{figure_polyhedron}-c).

The mentioned relaxation is valid for the following reason. 
Take two edges meeting at the vertex $\b{x}_i$. If the two edges belong to the same polyhedron face, the relaxation is exact. Consider the case where the two edges do not belong to the same face. These two edges are connected with a set of polyhedron faces. If we tweak one of the two edges to increase/decrease the angle between them, the angle of that edge with its neighbor edge on the same face also increases/decreases. Therefore, the changes in the additional angles of relaxation are consistent with the changes of the angles between the edges sharing the same faces.

\subsection{Curvature Calculation in the RKHS}\label{sec:kernel-curvature}

The pattern of curvature of data points might not be linear. Therefore, we use the \textit{kernel curvature} to work on data in the RKHS \cite{ghojogh2020anomaly}.
In kernel curvature calculation, the two stages of finding $k$-NN and calculating the curvature score are performed in RKHS. Let $\b{\phi}: \mathcal{X} \rightarrow \mathcal{H}$ be the pulling function mapping the data $\b{x} \in \mathcal{X}$ to the RKHS $\mathcal{H}$. In other words, $\b{x} \mapsto \b{\phi}(\b{x})$. Let $t$ denote the dimensionality of the RKHS, i.e., $\b{\phi}(\b{x}) \in \mathbb{R}^t$ while $\b{x} \in \mathbb{R}^d$. Note that we usually have $t \gg d$.
The kernel over two vectors $\b{x}_1$ and $\b{x}_2$ is the inner product of their pulled data \cite{hofmann2008kernel,ghojogh2023background}: 
\begin{align}
\mathbb{R} \ni k(\b{x}_1, \b{x}_2) := \b{\phi}(\b{x}_1)^\top \b{\phi}(\b{x}_2).
\end{align}
The Euclidean distance in the RKHS is \cite{scholkopf2001kernel}: 
\begin{align}
||\b{\phi}(\b{x}_i) - \b{\phi}(\b{x}_j)||_2 = \sqrt{k(\b{x}_i, \b{x}_i) -2 k(\b{x}_i, \b{x}_j) + k(\b{x}_j, \b{x}_j)}.
\end{align}
Using this distance, we find the $k$-NN of the dataset in the RKHS. 

After finding $k$-NN in the RKHS, we calculate the score in the RKHS. We pull the vectors $\breve{\b{x}}_a$ and $\breve{\b{x}}_b$ to the RKHS so $\breve{\b{x}}_a^\top \breve{\b{x}}_b$ is changed to $k(\breve{\b{x}}_a, \breve{\b{x}}_b) = \b{\phi}(\breve{\b{x}}_a)^\top \b{\phi}(\breve{\b{x}}_b)$. 
Let $\b{K}_i \in \mathbb{R}^{k \times k}$ denote the kernel matrix of neighbors of $\b{x}_i$ whose $(a,b)$-th element is $k(\breve{\b{x}}_a, \breve{\b{x}}_b)$. 
The vectors in Eq. (\ref{equation_curvature_score}) are normalized. In the RKHS, this is equivalent to normalizing the kernel \cite{ah2010normalized,ghojogh2023background}:
\begin{align}\label{equation_normalized_kernel}
k'(\breve{\b{x}}_a, \breve{\b{x}}_b) := \frac{k(\breve{\b{x}}_a, \breve{\b{x}}_b)}{\sqrt{k(\breve{\b{x}}_a, \breve{\b{x}}_a)\, k(\breve{\b{x}}_b, \breve{\b{x}}_b)}}.
\end{align}
If $\b{K}'_i \in \mathbb{R}^{k \times k}$ denotes the normalized kernel $\b{K}_i$, the kernel curvature score in the RKHS is:
\begin{align}\label{equation_kernel_curvature_score}
c(\b{x}_i) := \sum_{a=1}^{k-1} \sum_{b=a+1}^k \b{K}'_{i,ab},
\end{align}
where $\b{K}'_{i,ab}$ denotes the $(a,b)$-th element of the normalized kernel $\b{K}'_i$.

\section{CurvSSL and Kernel CurvSSL}   

\subsection{Network and Data Settings}

The neural network for self-supervised learning contains an encoder $f_\theta$ followed by a projection head $g$.
Let $\mathcal{X}\!\subset\!\mathbb{R}^d$ be the input space, $f_\theta:\mathcal{X}\!\to\!\mathbb{R}^{d_h}$ an encoder, and $g:\mathbb{R}^{d_h}\!\to\!\mathbb{R}^{d_z}$ a projection head.
Suppose $\mathcal{T}(\b{x})$ denotes the distribution of training data.
For every training data instance, we draw two stochastic augmentations $(\b{x},\b{x}')\sim\mathcal{T}(\b{x})$ and pass them through the encoder and the projection head:
\begin{equation}
\begin{aligned}
&\b{h}=f_\theta(\b{x}) \in \mathbb{R}^{d_h},\quad \b{z}=g(\b{h}) \in \mathbb{R}^{d_z},\\
&\b{h}'=f_\theta(\b{x}') \in \mathbb{R}^{d_h}, \quad \b{z}'=g(\b{h}') \in \mathbb{R}^{d_z}.
\end{aligned}
\end{equation}
Every mini-batch, with size $b$, is $\{(\b{z}_i,\b{z}'_i)\}_{i=1}^b$. 

\subsection{Loss Function}

We now describe the proposed self-supervised objective. As before, let
\(\{(\b{z}_i,\b{z}'_i)\}_{i=1}^b\) denote the projected embeddings of two augmentations of a mini-batch of size \(b\), where
\(\b{z}_i = g(f_\theta(\b{x}_i))\) and \(\b{z}'_i = g(f_\theta(\b{x}'_i)) \in \mathbb{R}^{d_z}\).
Our loss has two components:
(i) a redundancy-reduction term on the embedding coordinates, in the spirit of Barlow Twins,
and (ii) a curvature-based term that aligns and decorrelates curvature patterns across the batch.

\subsubsection{Redundancy reduction in embedding space}
We first normalize the projected embeddings per feature dimension:
\begin{equation}
\tilde{\b{z}}_i = \frac{\b{z}_i - \b{\mu}_z}{\b{\sigma}_z + \varepsilon}, 
\qquad
\tilde{\b{z}}'_i = \frac{\b{z}'_i - \b{\mu}'_z}{\b{\sigma}'_z + \varepsilon},
\end{equation}
where the division is element-wise, \(\b{\mu}_z, \b{\sigma}_z \in \mathbb{R}^{d_z}\) are the batch-wise mean and standard deviation of \(\{\b{z}_i\}_{i=1}^b\), and similarly for \(\b{\mu}'_z, \b{\sigma}'_z\) and \(\{\b{z}'_i\}_{i=1}^b\);
\(\varepsilon>0\) is a small constant for numerical stability.
We then form the cross-correlation matrix:
\begin{equation}
\b{C} \;\in\; \mathbb{R}^{d_z \times d_z}, 
\qquad
\b{C}_{uv} := \frac{1}{b} \sum_{i=1}^b \tilde{z}_{i,u}\,\tilde{z}'_{i,v},
\end{equation}
where $\b{C}_{uv}$ is the $(u,v)$-th element of $\b{C}$ and \(\tilde{z}_{i,u}\) denotes the \(u\)-th component of \(\tilde{\b{z}}_i\).
Following Barlow Twins \cite{zbontar2021barlow}, we enforce that the diagonal entries of \(\b{C}\) are close to~\(1\) (strong agreement between views in each feature) while off-diagonal entries are close to~\(0\) (low redundancy between different features):
\begin{equation}
\label{eq:barlow-loss}
\mathcal{L}_{\text{emb}}
:= \sum_{u=1}^{d_z} (\b{C}_{uu} - 1)^2
   \;+\;
   \lambda_{\text{emb}} \sum_{\substack{u,v=1 \\ u \neq v}}^{d_z} \b{C}_{uv}^2 ,
\end{equation}
where \(\lambda_{\text{emb}}>0\) controls the strength of the off-diagonal penalty.

\subsubsection{Curvature-based regularization}
In addition to redundancy reduction at the coordinate level, we regularize the \emph{local geometry} of the learned manifold via curvature.
For each embedding \(\b{z}_i\), we compute a discrete curvature score \(c(\b{z}_i)\) by treating \(\b{z}_i\) as a vertex of a hypothetical polyhedron whose faces are spanned by its \(k\)-nearest neighbors in the embedding space. Let \(\{\b{z}_{i,a}\}_{a=1}^k\) denote these neighbors, and define edge vectors
\(\breve{\b{z}}_{i,a} := \b{z}_{i,a} - \b{z}_i\).
Normalizing these edges onto the unit hypersphere and aggregating the pairwise cosine similarities between neighbor directions yields the curvature score (see Eq. (\ref{equation_curvature_score})):
\begin{equation}
\label{eq:curv-score-emb}
c(\b{z}_i)
:= \sum_{a=1}^{k-1} \sum_{b=a+1}^k
\frac{\breve{\b{z}}_{i,a}^\top \breve{\b{z}}_{i,b}}
{\|\breve{\b{z}}_{i,a}\|_2 \,\|\breve{\b{z}}_{i,b}\|_2},
\end{equation}
which measures how sharply the local neighborhood around \(\b{z}_i\) bends.
The kernel curvature score is (see Eq. (\ref{equation_kernel_curvature_score})):
\begin{align}\label{eq:kernel-curv-score-emb}
c(\b{z}_i) := \sum_{a=1}^{k-1} \sum_{b=a+1}^k \b{K}'_{i,ab},
\end{align}
where $\b{K}'_{i,ab}$ is the $(a,b)$-th element of the normalized kernel kernel $\b{K}'_i$ between $\breve{\b{z}}_{i,a}$ and $\breve{\b{z}}_{i,b}$.

Eqs. (\ref{eq:curv-score-emb}) and (\ref{eq:kernel-curv-score-emb}) can be used for curvature scores in CurvSSL and kernel CurvSSL loss functions, respectively. 
We compute analogous curvature scores \(c(\b{z}'_i)\) for the second view.

Stacking the curvature scores into vectors
\(\b{c} = [c(\b{z}_1),\dots,c(\b{z}_b)]^\top\) and
\(\b{c}' = [c(\b{z}'_1),\dots,c(\b{z}'_b)]^\top \in \mathbb{R}^b\),
we first normalize them across the batch:
\begin{equation}
\tilde{\b{c}} = \frac{\b{c} - \mu_c \b{1}}{\sigma_c + \varepsilon},
\qquad
\tilde{\b{c}}' = \frac{\b{c}' - \mu'_c \b{1}}{\sigma'_c + \varepsilon},
\end{equation}
where \(\mu_c, \sigma_c \in \mathbb{R}\) are the mean and standard deviation of \(\b{c}\),
\(\mu'_c, \sigma'_c\) are those of \(\b{c}'\), the \(\b{1} \in \mathbb{R}^b\) is the all-ones vector,
and \(\varepsilon>0\) is again a small constant.
We then form a curvature-derived matrix:
\begin{equation}
\label{eq:curv-matrix}
\b{M} \;\in\; \mathbb{R}^{b \times b},
\qquad
\b{M}_{ij} := \frac{1}{b}\,\tilde{c}_i \,\tilde{c}'_j,
\end{equation}
where $\b{M}_{ij}$ denotes the $(i,j)$-th element of $\b{M}$, which plays an analogous role to the cross-correlation matrix \(\b{C}\), but now at the \emph{sample} level in terms of curvature.
We encourage the curvature of matched augmentations to agree (diagonal entries of \(\b{M}\) close to~\(1\)) and the curvature patterns of different samples to be decorrelated (off-diagonals close to~\(0\)):
\begin{equation}
\label{eq:curv-loss}
\mathcal{L}_{\text{curv}}
:= \sum_{i=1}^{b} (\b{M}_{ii} - 1)^2
   \;+\;
   \lambda_{\text{curv}} \sum_{\substack{i,j=1 \\ i \neq j}}^{b} \b{M}_{ij}^2 ,
\end{equation}
where \(\lambda_{\text{curv}}>0\) controls the strength of curvature-based redundancy reduction.

\subsubsection{Total objective and kernel extension}
Our final self-supervised loss is a weighted sum of the embedding-level and curvature-level terms:
\begin{equation}
\label{eq:total-loss}
\mathcal{L}
:= \mathcal{L}_{\text{emb}}
   \;+\;
   \alpha_{\text{curv}}\,\mathcal{L}_{\text{curv}},
\end{equation}
where \(\alpha_{\text{curv}}>0\) balances the influence of curvature regularization.
In the Euclidean case, i.e., CurvSSL, \(c(\cdot)\) is given by Eq. \eqref{eq:curv-score-emb}.
In the kernel curvature variant, i.e., kernel CurvSSL, Eq. (\ref{eq:kernel-curv-score-emb}) is used for \(c(\cdot)\).

The proposed objective enforces view invariance and redundancy reduction at the level of embedding coordinates, while simultaneously shaping the local manifold geometry through curvature alignment and curvature-based decorrelation across the batch.

\section{Experiments}

We empirically evaluate the proposed curvature-regularized self-supervised learning on two standard benchmarks, MNIST \cite{lecun1998gradient} and CIFAR-10 \cite{krizhevsky2009learning}, using a ResNet backbone \cite{he2016deep} and a two-stage protocol: (i) self-supervised pretraining with the proposed CurvSSL and kernel CurvSSL objectives, and (ii) frozen-encoder linear evaluation. In addition, we visualize the learned representations with UMAP~\cite{mcinnes2018umap} to inspect the geometry induced by curvature regularization.

\subsection{Experimental Setup}

\paragraph{Datasets.}
We consider MNIST \cite{lecun1998gradient} and CIFAR-10 \cite{krizhevsky2009learning} as two representative image datasets of increasing difficulty.
MNIST consists of grayscale handwritten digits (10 classes), while CIFAR-10 contains natural RGB images with more complex intra-class variability.
For SSL pretraining, we use only the training split of each dataset.
For linear evaluation, we use the standard training and test splits.


\paragraph{Network and training details.}
For both datasets, we adopt a ResNet-18 \cite{he2016deep} encoder $f_\theta$ followed by a two-layer MLP projection head $g$ that maps encoder features to a $d_z$-dimensional projection space.
The self-supervised model is trained using the curvature-regularized loss~\eqref{eq:total-loss}, where the embedding-level redundancy reduction $\mathcal{L}_{\text{emb}}$ is instantiated as a Barlow Twins objective~\eqref{eq:barlow-loss}, and the curvature-level term $\mathcal{L}_{\text{curv}}$ uses the discrete curvature score~\eqref{eq:curv-score-emb} or (\ref{eq:kernel-curv-score-emb}) with $k$-nearest neighbors in the projected space. We use $d_z = 128$, $k = 10$ neighbors, a mini-batch size of $256$, and train the SSL model for $100$ epochs for MNIST and $500$ epochs for CIFAR-10 using Adam optimizer \cite{kingma2014adam} with learning rate $10^{-3}$ and weight decay $10^{-4}$.
The curvature and embedding weights $(\lambda_{\text{emb}}, \lambda_{\text{curv}}, \alpha_{\text{curv}})$ are selected once and reused across datasets. In all experiments, we simply set $\lambda_{\text{emb}} = \lambda_{\text{curv}} = \alpha_{\text{curv}} = 1$.
For kernel CurvAlign, radial basis function (RBF) kernel function was employed.


\paragraph{Data augmentations.}
For MNIST, we follow common practice for SSL on digit images, applying random resized crops, small rotations, and grayscale-to-RGB conversion, followed by per-channel normalization.
For CIFAR-10, we adopt a standard augmentation pipeline with random resized crops, horizontal flips, color jitter, random grayscale, and per-channel normalization.
Two independent augmented views are sampled for each image in a mini-batch and passed through the shared encoder--projector.

\subsection{Linear Evaluation}

To assess the quality of the learned representations, we perform linear evaluation following the standard protocol.
After SSL pretraining, we freeze the encoder $f_\theta$ and discard the projection head $g$.
A small classifier consisting of a linear layer with one hidden layer and batch normalization (as described in Section~3) is trained on top of the frozen encoder features using cross-entropy loss.
Only the classifier parameters are updated; the encoder weights remain fixed.

We train the linear classifier for a fixed number of epochs (e.g., $50$) with SGD and report top-1 test accuracy on MNIST and CIFAR-10.
The results are summarized in Table~\ref{tab:linear-eval}.
Overall, CurvSSL and kernel CurvSSL achieve competitive linear probe performance on both datasets, indicating that enforcing both redundancy reduction and curvature consistency produces representations that transfer well to supervised classification.
On CIFAR-10, which is more challenging, we observe that the curvature term does not prevent the model from learning discriminative features and can improve class separation compared to using redundancy reduction alone.
Moreover, as expected, kernel CurvSSL performs better than CurvSSL because of handling nonlinearities better through RKHS.

\begin{table}[t]
    \centering
    \small 
    \caption{Linear evaluation accuracy (\%) on MNIST and CIFAR-10 using a frozen ResNet-18 encoder pretrained with the SSL objectives.}
    \label{tab:linear-eval}
    \begin{tabular}{lcc}
        \hline
        \\
        Method & MNIST & CIFAR-10 \\
        \hline
        \\
        VicReg \cite{bardes2022vicreg} & \textit{95.9} & \textit{74.5} \\
        Barlow Twins \cite{zbontar2021barlow}  & \textit{94.9} & \textit{73.6} \\
        CurvSSL (ours) & \textbf{\textit{97.9}} & \textbf{\textit{75.1}} \\
        Kernel CurvSSL (ours) & \textbf{\textit{98.4}} & \textbf{\textit{76.5}} \\
        \hline
    \end{tabular}
\end{table}

\begin{figure}[t]
  \centering

  \begin{minipage}{\linewidth}
    \centering
    \includegraphics[width=0.8\linewidth]{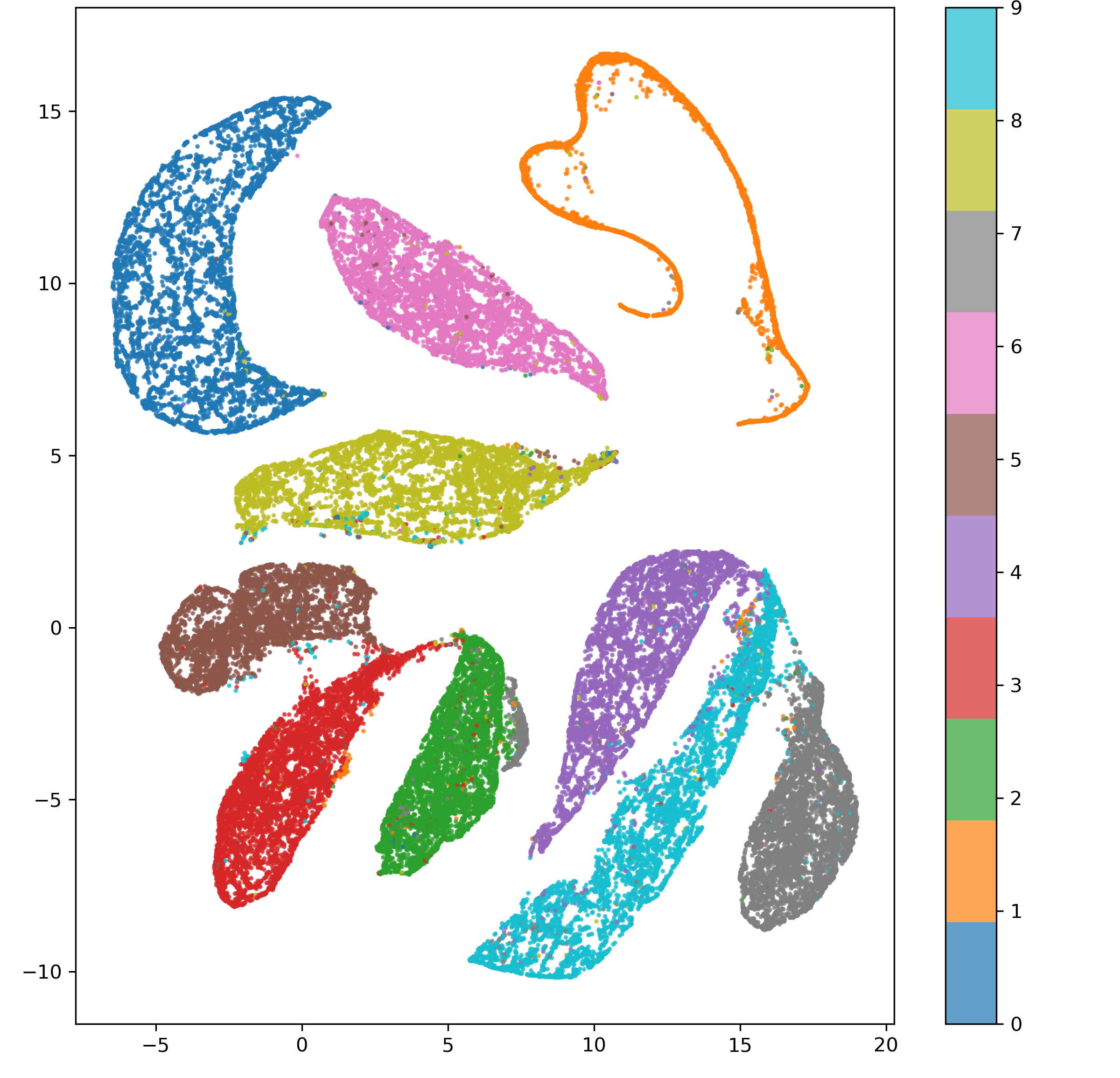} 
    \\[4pt]
    {\small\textbf{(a)} CurvSSL (Euclidean).}
  \end{minipage}

  \vspace{6pt} 

  \begin{minipage}{\linewidth}
    \centering
    \includegraphics[width=0.8\linewidth]{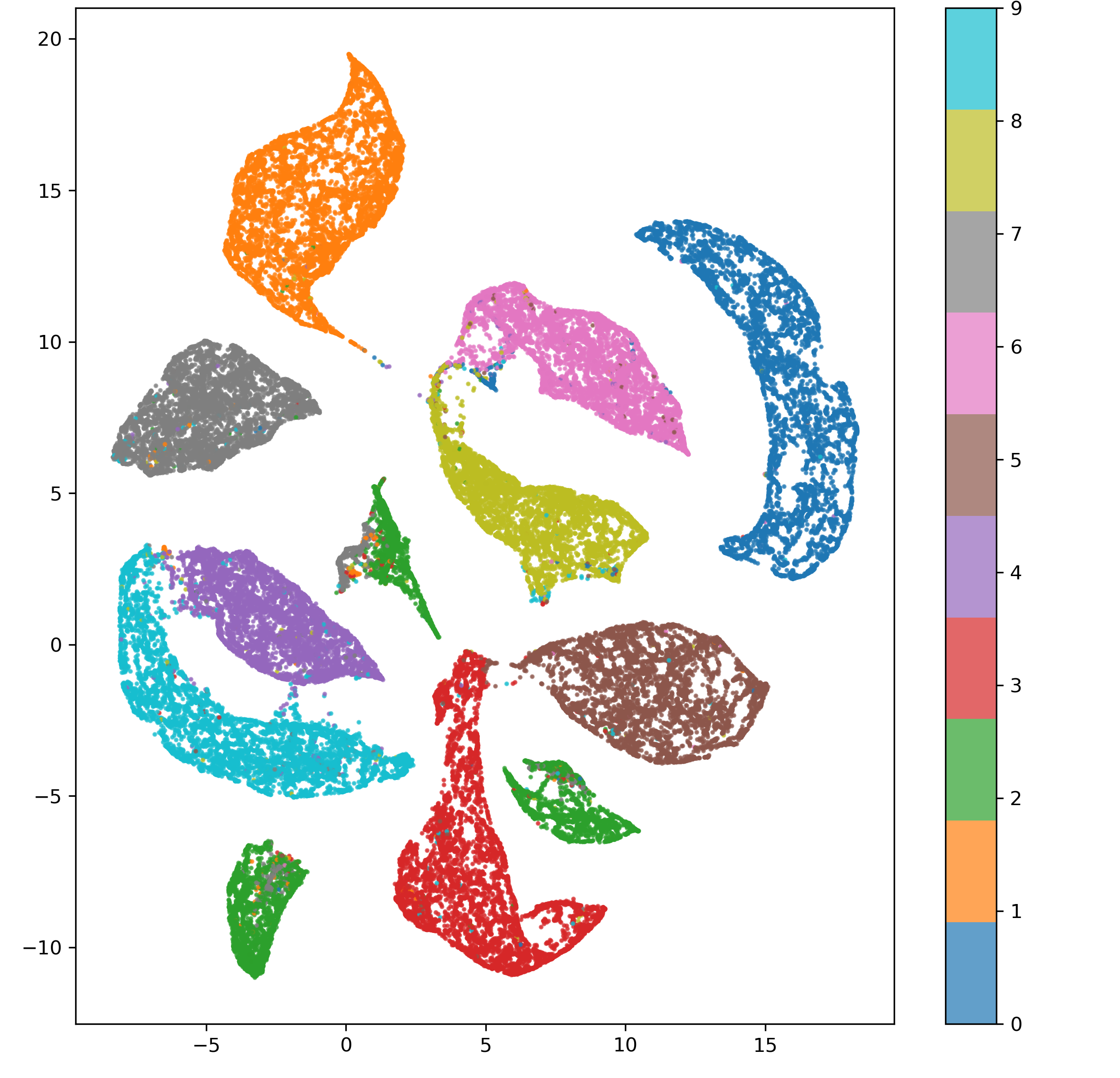} 
    \\[4pt]
    {\small\textbf{(b)} Kernel CurvSSL.}
  \end{minipage}

  \caption{UMAP visualization of encoder features on MNIST after curvature-regularized SSL. Points are colored by ground-truth digit class.}
  \label{fig:umap-mnist}
\end{figure}

\subsection{UMAP Visualization of Learned Representations}

Beyond scalar accuracy, we study the geometry of the learned representations via UMAP embeddings.
For each dataset, we extract features from the frozen encoder on a held-out split (train or test) and project them to two dimensions using UMAP with a fixed configuration (e.g., $n_{\text{neighbors}}=15$, $\text{min\_dist}=0.1$).
We visualize either the encoder features $\b{h}$ or the projected features $\b{z}$, coloring points by ground-truth class labels.

Figure~\ref{fig:umap-mnist} compares Euclidean CurvSSL and its kernel variant on MNIST.
Clusters corresponding to different digits are well separated, with relatively smooth transitions between nearby classes (e.g., visually similar digits such as `3' and `5').
The curvature regularization encourages local neighborhoods to be geometrically consistent across augmentations, which manifests as tighter and more coherent class clusters in the UMAP plot.

\begin{figure}[t]
  \centering

  \begin{minipage}{\linewidth}
    \centering
    \includegraphics[width=0.8\linewidth]{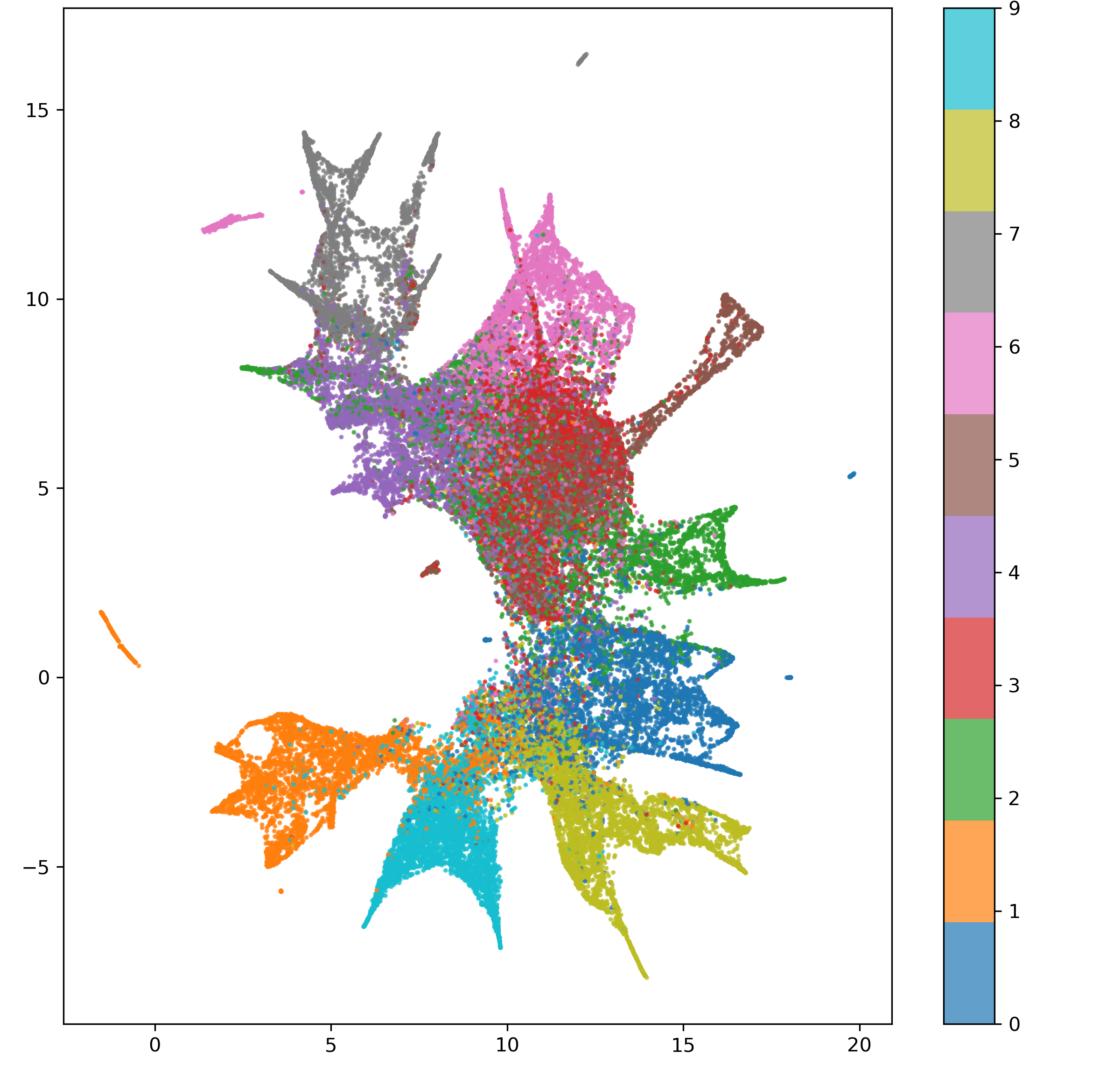} 
    \\[4pt]
    {\small\textbf{(a)} CurvSSL (Euclidean).}
  \end{minipage}

  \vspace{6pt} 

  \begin{minipage}{\linewidth}
    \centering
    \includegraphics[width=0.8\linewidth]{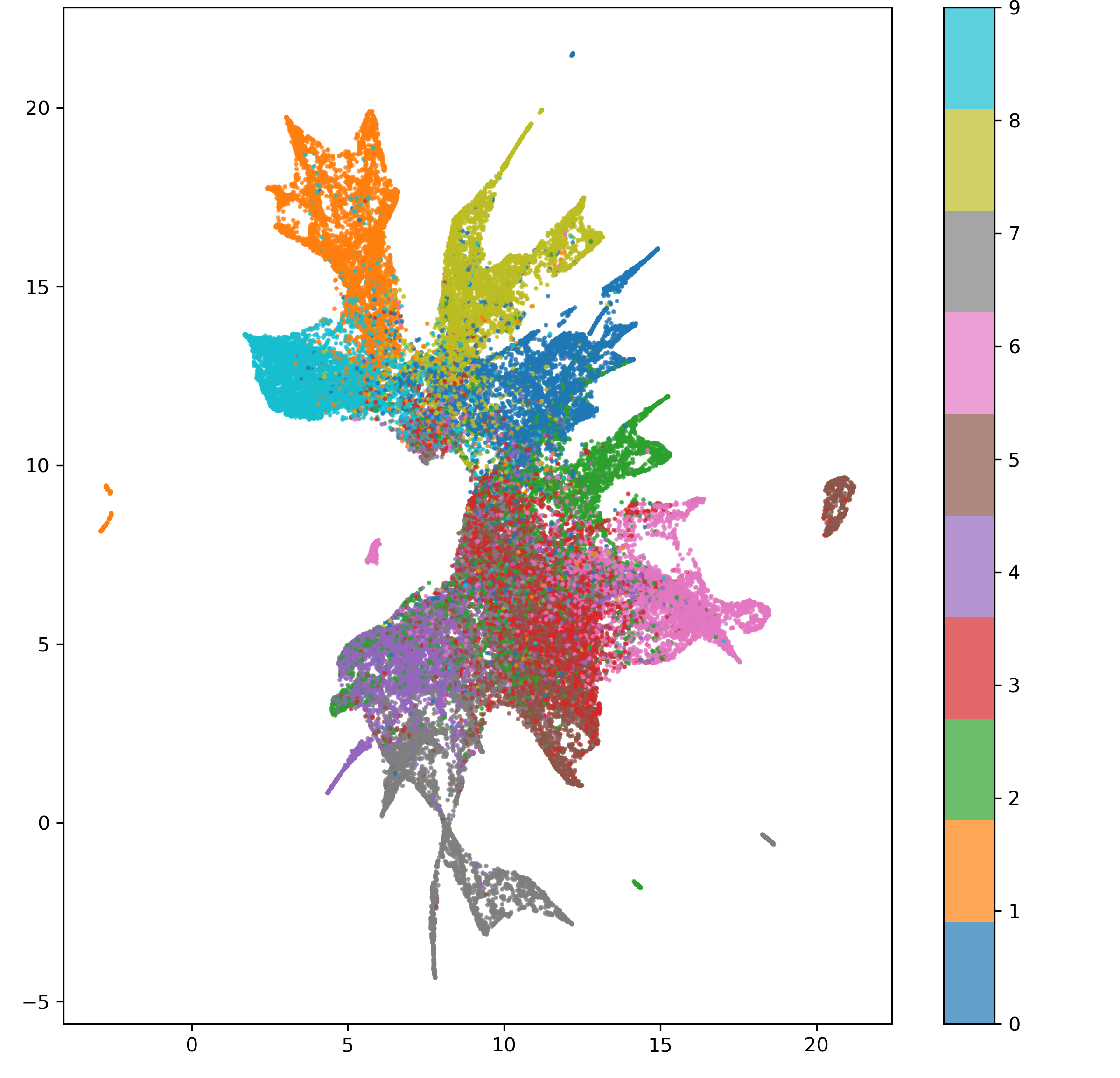} 
    \\[4pt]
    {\small\textbf{(b)} Kernel CurvSSL.}
  \end{minipage}

  \caption{UMAP visualization of encoder features on CIFAR-10 after curvature-regularized SSL. Points are colored by ground-truth class.}
  \label{fig:umap-cifar}
\end{figure}

As depicted in Fig.~\ref{fig:umap-cifar} for CIFAR-10, both CurvSSL and Kernel CurvSSL produce embeddings that form more complex structures, reflecting the higher intra-class variability of natural images.
Nonetheless, we observe that classes occupy distinct regions with meaningful local neighborhoods.

\section{Conclusion}

We proposed geometry-aware self-supervised objectives, named CurvSSL and kernel CurvSSL, that augment a Barlow Twins-style redundancy reduction loss with curvature-based regularizations. By treating each embedding as a vertex with a discrete curvature score computed from its $k$-nearest neighbors on the unit hypersphere, and coupling these scores across augmentations and samples via a curvature–Barlow loss, our method encourages both view invariance and consistency of local manifold geometry.

On MNIST and CIFAR-10, curvature-regularized SSL yields competitive linear evaluation accuracy and well-structured UMAP embeddings, suggesting that explicitly shaping local geometry complements standard invariance and redundancy-reduction terms. The method is simple to integrate into existing two-view pipelines and admits a kernel extension, making it a practical starting point for further geometric SSL work on larger datasets, architectures, and manifold-sensitive tasks such as semi-supervised learning and retrieval.

\bibliography{ref}
\bibliographystyle{icml2025}


\end{document}